\documentclass[conference]{IEEEtran}
\IEEEoverridecommandlockouts

\usepackage{cite}
\usepackage{subcaption}
\usepackage{amsmath,amssymb,amsfonts}
\usepackage{algorithmic}
\usepackage{graphicx}
\usepackage{textcomp}
\usepackage{xcolor}
\usepackage{wrapfig}

\usepackage{multirow}\usepackage[symbol]{footmisc}

\usepackage[colorlinks=true,pagebackref=false,citecolor=blue,bookmarks=false,hypertexnames=true]{hyperref}
\def\BibTeX{{\rm B\kern-.05em{\sc i\kern-.025em b}\kern-.08em
   T\kern-.1667em\lower.7ex\hbox{E}\kern-.125emX}}
\begin{document}

\title{SpikiLi: A Spiking Simulation of LiDAR based Real-time Object Detection for Autonomous Driving}

\author{Sambit Mohapatra*$^{1}$, 
Thomas Mesquida*$^{2}$, 
Mona Hodaei$^{1}$, 
Senthil Yogamani$^{3}$, 
Heinrich Gotzig$^{1}$,  
Patrick M\"ader$^{4}$ \\
{\normalsize 
*Equal contribution \hspace{0.3cm} 
$^{1}$Valeo Germany\hspace{0.3cm} 
$^{2}$CEA-List, France \hspace{0.3cm}
$^{3}$Valeo Ireland \hspace{0.3cm} 
$^{4}$TU Ilmenau, Germany 
}
}


\maketitle
\begin{abstract}
Spiking Neural Networks are a recent and new neural network design approach that promises tremendous improvements in power efficiency, computation efficiency, and processing latency. They do so by using asynchronous spike-based data flow, event-based signal generation, processing, and modifying the neuron model to resemble biological neurons closely. While some initial works have shown significant initial evidence of applicability to common deep learning tasks, their applications in complex real-world tasks has been relatively low. In this work, we first illustrate the applicability of spiking neural networks to a complex deep learning task namely Lidar based 3D object detection for automated driving. Secondly, we make a step-by-step demonstration of simulating spiking behavior using a pre-trained convolutional neural network. We closely model essential aspects of spiking neural networks in simulation and achieve equivalent run-time and accuracy on a GPU. When the model is realized on a neuromorphic hardware, we expect to have significantly improved power efficiency.
\end{abstract}
\section{Introduction}

Recently, there has been a lot of work on perception related to various tasks such as object detection \cite{rashed2021generalized, sekkat2022synwoodscape}, soiling detection \cite{uricar2021let, das2020tiledsoilingnet}, motion segmentation \cite{yahiaoui2019fisheyemodnet, rashed2019motion}, road edge detection \cite{dahal2021roadedgenet, dahal2021online}, weather classification \cite{dhananjaya2021weather}, depth prediction \cite{thesis_varun, kumar2021svdistnet, kumar2018monocular, kumar2020fisheyedistancenet, kumar2021syndistnet, kumar2021fisheyedistancenet++, kumar2020syndistnet}, SLAM \cite{gallagher2021hybrid, cheke2022fisheyepixpro} and in general for multi-task outputs \cite{sistu2019real, phd_varun, kumar2021omnidet, klingner2022detecting, chennupati2019auxnet, sobh2021adversarial, kia_2021} based on Convolutional Neural Networks (CNNs). However CNNs are power hungry and it is difficult to deploy all the above tasks in a low power embedded system. Spiking Neural Networks (SNNs) are the most recent approach to designing deep neural networks that are closely modeled after signal processing in biological neural circuits such as the brain. The motivation here is to mimic biological neural networks to meet the low power and compute requirements necessary to deploy state-of-the-art neural networks in real-world applications on embedded platforms. An SNN typically has the following key properties:
\begin{itemize}
    \item \textbf{Spike coding:} Data is encoded into equal amplitude pulses (or digital signals), where information is carried by the frequency (rate-based) \cite{adrian1926impulses} of pulses or time of pulse emission (temporal) \cite{park2019fast} \cite{butts2007temporal}. It directly impacts computation requirements since the signals can now be represented with just 1s and 0s in the digital domain.
    \item \textbf{Event-based processing:} SNNs process events rather than signals directly (i.e., changes in perceived signals). As a result, minimal variations, which most often do not contain essential and valuable information, are not processed, leading to energy efficiency.
    \item \textbf{Asynchronous operation:} Biological neurons do not have a global clock and process events asynchronously. SNNs try to emulate this behavior which leads to very low latency. Unlike frame-based CNNs, SNNs process events, and events are generated asynchronously. Therefore, only those regions of the input data are processed, which generate events. However, achieving proper asynchronous operation is only feasible when implemented on neuromorphic hardware that allows parallel operation of different input regions and networks underneath.
    \item \textbf{Biologically plausible neuron models:} SNNs use neuron models such as Leaky Integrate and Fire (LIF) \cite{maass1997networks} and Integrate and Fire (IF)~\cite{eason1955certain} that are close mathematical approximations of biological neuron models. They help maintain the sparsity of activations throughout the network, leading to lower power consumption. 
\end{itemize}
    SNNs can either be independently designed and trained using unsupervised training schemes like Spike Time Dependent Plasticity (STDP)~\cite{song2000competitive} or a CNN can be trained using conventional back-propagation \cite{lecun1988theoretical}. Then the trained network is converted layer by layer to an equivalent spiking model. This approach has been the most successful \cite{sengupta2019going} in terms of accuracy and precision, and we follow this approach in this work. Our key contributions are:
\begin{itemize}
    \item We demonstrate block-wise conversion of a CNN to SNN using floating-point weights from a pre-trained CNN and retrained using 4-bit signed integer quantized weights for spike-based input.
    
    \item We design and demonstrate an efficient spike coding scheme using a learnable parameter that can convert conventional point cloud data sets into spike trains.
    \item We demonstrate that an SNN can achieve comparable precision for a complex task on the KITTI \cite{geiger2013vision} dataset.
\end{itemize}
\begin{figure*}[t]
        \centering
        \captionsetup{justification=centering, singlelinecheck=false, font=small, belowskip=-12pt}
        \includegraphics[width=\textwidth]{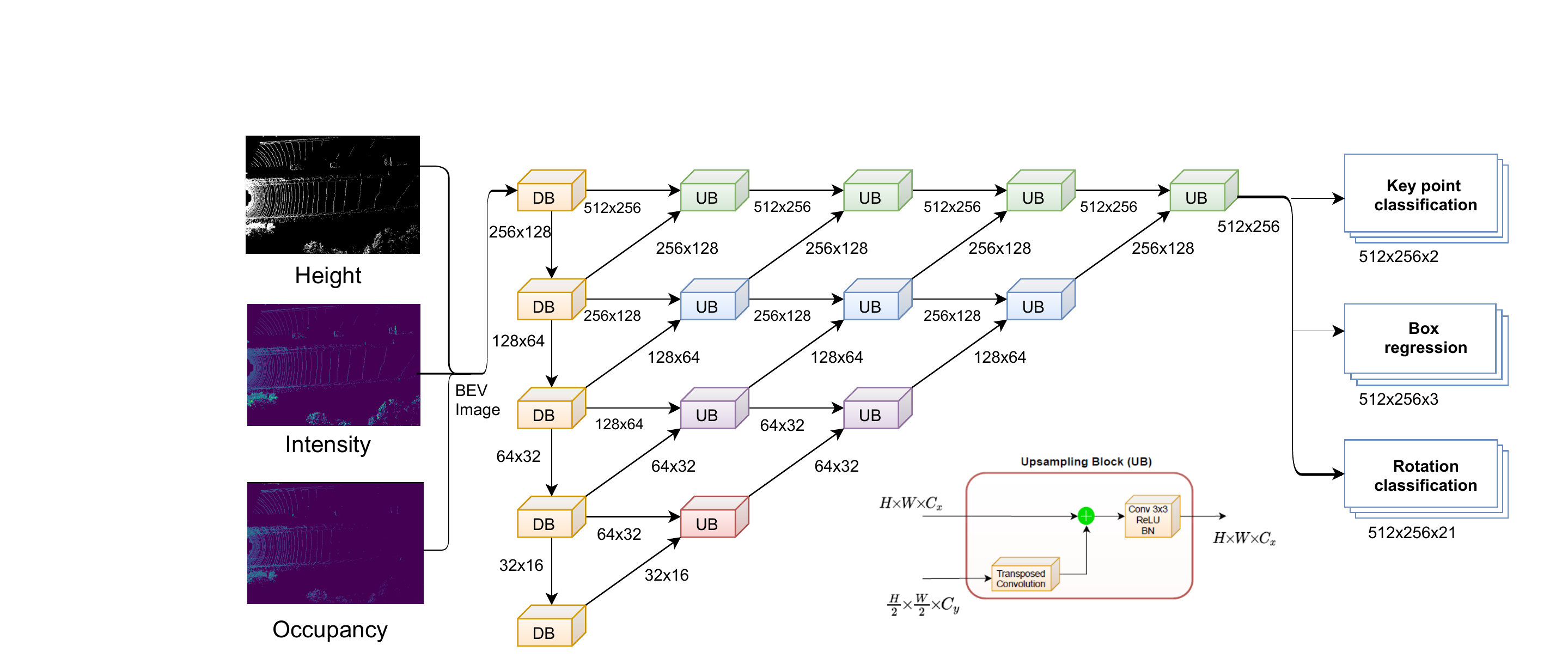}
         \includegraphics[width=\textwidth]{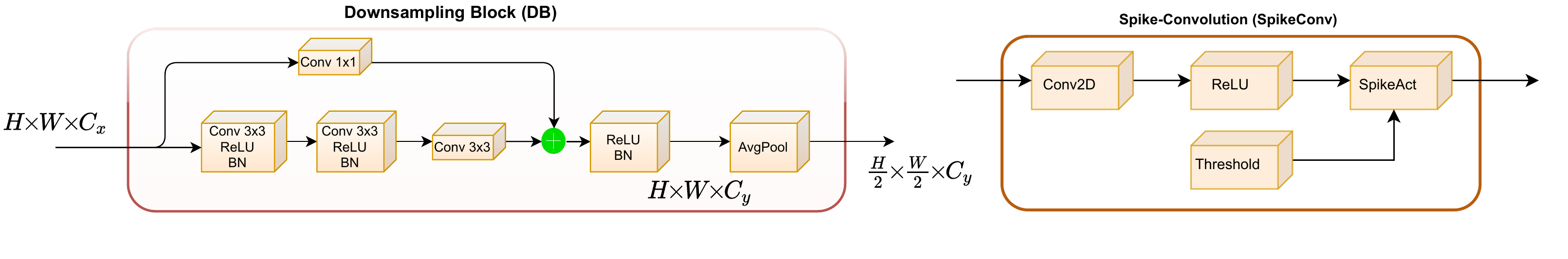}
         \centering
        \caption{\bf Our BEVDetNet~\cite{mohapatra2021bevdetnet} Network architecture adapted by modified convolution with spiking activation (SpikeConv block).}
        \label{fig:Figure_overall_nw_bd}
\end{figure*}
\section{Related work}

Although Spiking Neural Networks are a relatively new generation of neural networks, there have been some investigations into employing them in applications other than neuromorphic ones. SNNs have been studied less than CNNs; however, there are already fully functional converted SNNs that can deal with modern vision tasks~\cite{sengupta2019going} such as image classification~\cite{mostafa2017supervised}\cite{rueckauer2017conversion} and object detection~\cite{kheradpisheh2018stdp}\cite{kim2020spiking}.

Diehl et al.\cite{diehl2015fast} have focused on the MNIST dataset~\cite{lecun1998gradient} for handwritten digits, using a conversion technique to convert a CNN into SNN. In another experiment \cite{hunsberger2015spiking} used a Leaky Integrate and Fire (LIF) neuron model with a smoothened response, converting CNN to SNN for object recognition application on the CIFAR-10 dataset. While these methods show impressive accuracy and excellent energy-efficient potential, such direct conversion of pre-trained networks without fine-tuning is only viable for small networks and small datasets, such as image classification on MNIST and CIFAR. For deeper architectures, the loss inaccuracy would be a concern. More recently, \cite{rathi2020enabling} proposed a spike time-dependent back-propagation (STDB) method that allows using supervised learning efficiently with SNNs. Weights from a pre-trained neural network are used as initializers, and then the network is retrained using STDB. Impressive results have been demonstrated on the CIFAR-10 and ImageNet~\cite{deng2009imagenet} datasets.

Because of the nature of SNNs which are based on the non-differentiability of spiking activities, training large-scale complex networks has always been difficult, and SNNs have been limited to simple tasks and small datasets~\cite{gehrig2020event} during the past years. However, designing SNNs for real-world robotics and computer vision applications has recently reached advancements because of new spike learning mechanisms~\cite{zenke2018superspike}\cite{shrestha2018slayer}. SNNs have been used over traditional frame-based images in order to perform object detection and recognition \cite{cannici2019asynchronous}\cite{zhang2019tdsnn}\cite{lee2016training}. SNN approaches have been applied in order to do segmentation tasks. Studies \cite{meftah2010segmentation}\cite{zheng2019image} investigated region segmentation without including semantic information, for example, class information. In another study, SNN has shown robust and energy-efficient performance in the semantic segmentation task~\cite{kim2021beyond}. The combination of this advancement with neuromorphic processors like IBM’s TrueNorth \cite{akopyan2015truenorth} and Intel’s® Loihi \cite{davies2018loihi} accompanying neuromorphic sensors such as event-based cameras (e.g, DVS) \cite{lichtsteiner2008128} and ATIS \cite{posch2010qvga} brings the ability to work better on real-world problems. In~\cite{kirkland2020spikeseg}, a SNN trained with STDP performed great in the semantic event-based image segmentation task. Also, some advancements have been made in achieving unsupervised learning of SNNs for event-based optical flow estimation\cite{hagenaars2021self}\cite{paredes2019unsupervised}. A number of other studies have been conducted to use SNNs in object detection and recognition, using event-based LiDAR and DVS data~\cite{wu2019direct}\cite{zhou2018object}.\par
\section{Proposed Method}

In this work, we propose a method to convert an optimized and pre-trained CNN \cite{mohapatra2021bevdetnet} into spiking equivalent by adding wrapper layers around the convolutional units to simulate Integrate and Fire (IF) spiking behavior. As seen in Fig. \ref{fig:Figure_overall_nw_bd}, the base CNN architecture takes in the bird's eye view (BEV) images of a 3D point cloud and produces three outputs, each at the same spatial resolution as the input. The keypoint head classifies certain pixels as key points where a key point is the object's center. Corresponding to each predicted keypoint, the box regression head predicts the height, width, and length of a box that would enclose the object corresponding to the keypoint.
Similarly, for each predicted key point, the rotation classification head predicts the box's rotation as a binned classification task. In the simulation, actual spikes are represented by integers. We use the pre-trained weights as a starting point and then re-train the network using standard back-propagation while quantizing the weights in each pass. The addition of the IF model and spike activations propagates sparsity of activations throughout the network. Several of the modified convolutional layers produce zeros as outputs due to the spike activation threshold. This directly translates to no switching of electronic switches and power consumption in neuromorphic hardware. To keep things simple for the simulation, we do not accumulate spikes over time and process them instantaneously. Following this, all our data encoding and decoding follow a rate-based approach rather than temporal.\par
\begin{table}[t]
\centering
\captionsetup{singlelinecheck=false, font=footnotesize}
\centering
\caption{\centering{ \bf Average Precision (AP) and latency comparison between the original CNN and simulated SNN on KITTI validation dataset.}}
\label{tab:my-table}
\scalebox{0.8}{
\small
\setlength{\tabcolsep}{0.43em}
\begin{tabular}{|c|c|c|c|c|c|c|c|c|} 
\hline
\multicolumn{1}{|c|}{\textbf{\begin{tabular}[c]{@{}c@{}}Val\\ Split\end{tabular}}} & \multicolumn{1}{c|}{\textbf{\begin{tabular}[c]{@{}c@{}}BEVDetNet\\ Model\end{tabular}}}  & \multicolumn{3}{c|}{\textbf{\rule{0pt}{7pt}AP at IoU = 0.5}} & \multicolumn{3}{c|}{\textbf{\rule{0pt}{7pt}AP at IoU = 0.7}} & \multirow{2}{*}{\begin{tabular}[c]{@{}c@{}}\textbf{Infer.}\\\textbf{(ms)}\end{tabular}} \\ 
\cline{3-8}
 &  & \multicolumn{1}{c|}{\textbf{Easy}} & \multicolumn{1}{c|}{\textbf{Moderate}} & \textbf{Hard} & \multicolumn{1}{c|}{\textbf{Easy}} & \multicolumn{1}{c|}{\textbf{Moderate}} & \textbf{Hard} &  \\ 
\hline 
\multirow{2}{*}{50/50} & \rule{0pt}{7pt} CNN & 87.82 & 87.78 & 87.25 & 82.46 & 77.90 & 77.45 & 3 \\ 
\cline{2-9}\rule{0pt}{7pt}
 & SNN & 85.43 & 78.59 & 77.67 & 73.31 & 72.50 & 66.32 & 3 \\
\hline
\cline{2-9}{75/25}\rule{0pt}{7pt}
 & SNN & 86.37 & 78.65 & 78.76   & 82.71 & 75.49 & 68.68 & 3 \\
\hline
\end{tabular}
}
\vspace{-2em}
\end{table}
During training, the floating-point inputs are converted to spikes (represented by integers), a learnable parameter for rate-based spike generation. The forward pass, backward pass, and gradients are updated as usual using the PyTorch \cite{paszke2019pytorch} deep learning framework. After each batch, we quantize the weights to 4-bit signed integers and thresholds to 6-bit signed integers. The training process is continued this way till appropriate accuracy is achieved.\par
\subsection{Integrate and Fire neuron model and spike activation}

Integrate and Fire (IF) neuron model is probably the most simple and commonly used neuron model for the implementation of SNNs. It is a simplification of the Leaky Integrate and Fire (LIF) given by (\ref{eq1}):
\begin{align}\label{eq1}
    \tau_{m}\frac{dV_{mem}}{dt} = -V_{mem} + w * \theta_{t}
\end{align}
where \(V_{mem} : membrane \hspace{2pt}potential\), 

\(\theta_{t} : spike \hspace{2pt} at \hspace{2 pt}time \hspace{2pt}t\) and 

\(w : weight \hspace{2pt} of \hspace{2pt} connecting \hspace{2pt}synapse\).\par

We ignore the leakage part in the IF model, and at any time instant, the membrane potential of a neuron depends only on its previous membrane potential and the weighted sum of incoming spikes given (\ref{eq2}) for the discreet case:
\begin{align}\label{eq2}
    \Delta{V_{mem, t}} = V_{mem, t-1} + \sum_{k}^{K}{\theta_{k} * w_{k}}
\end{align}
where \(V_{mem, t} : membrane \hspace{2pt}potential \hspace{2pt} at \hspace{2pt}time \hspace{2pt} t\),

\(\theta_{k} : spike \hspace{2pt} on \hspace{2pt} k^{th} \hspace{2pt}neuron \hspace{2pt}at \hspace{2 pt}time \hspace{2pt}t\),

\(w : weight \hspace{2pt} of \hspace{2pt} connecting \hspace{2pt}synapse\) and

\(K: total \hspace{2pt} number \hspace{2pt} of \hspace{2pt}neurons\).\par
In this model, each neuron is characterized by a membrane potential: the neuron's present state, a threshold potential, a series of synapses bringing spikes into the neuron, and a series of synapses carrying spikes out of the neuron. Whenever the sum of pre-synaptic inputs weighted by the synaptic weights crosses the threshold value, the neuron emits a spike. In our simulation, each element of a convolution kernel can be thought of as the neuron connected to neurons in the layer before and after it. As can be seen in Fig. \ref{fig:Figure_overall_nw_bd}, each ReLU is followed by another activation block (SpikeAct) that receives a trainable threshold parameter and additional input. The SpikeAct converts membrane potential accumulated per inference to spike count to be transfered to the next layers. Its trainable 6b threshold controls the amount of spikes that has to be produced. We replace each of the convolution-based blocks (including transposed convolution and BatchNorm) of \cite{mohapatra2021bevdetnet} in the downsampling and upsampling blocks as seen in Fig. \ref{fig:Figure_overall_nw_bd} with this modified spike-convolutional block. In the backward pass, the gradients are copied and propagated backward from the SpikeAct.\par
\subsection{Input pre-processing}

As in \cite{mohapatra2021bevdetnet}, we convert each point cloud from the training dataset into a 2D BEV composed of three channels - height, intensity, and a binary mask. 
Each pixel of the BEV is then converted into a spike train. We use a rate-based spike coding technique. However, unlike the commonly used Poisson process-based rate coding, we use a learnable parameter to scale the floating-point input data to integer values. These integers in simulation correspond to a series of spikes in actual neuromorphic hardware.
The first layer of the network serves as a spike coder layer taking in conventional input data and producing spikes represented by integers.\par
\subsection{Retraining and fine tuning}

\begin{figure}[t]
\captionsetup{singlelinecheck=false, font=footnotesize, belowskip=-10pt}
\includegraphics[height=0.2\textheight]
{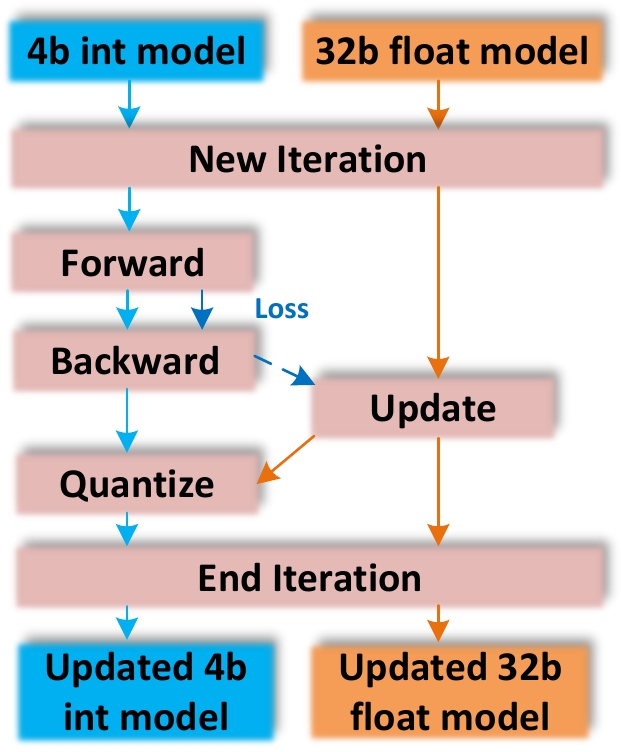}
\centering
\captionsetup{justification=centering} 
\caption{\bf Flowchart for training quantized SNN model.}
\centering
\label{fig:spikeConvTrain}
\vspace{-0.5em}
\end{figure}

Training quantized SNN from scratch or fine-tuning trained topologies requires two copies of the network parameters should be kept at all times: The model quantized to the desired precision and a full precision clone. The quantized version performs the forward and backward pass for meaningful gradient computation. The update is then performed on the full precision model. Updated full precision parameters are then quantized back to update the quantized model. Using this scheme, the weight updates within the heavily quantized model can be smoothed. Fig. \ref{fig:spikeConvTrain} provides an overview of the training pipeline. We use focal loss \cite{lin2017focal} for the key points and rotation heads and mean squared error loss for the box regression head. We perform training and validation on the KITTI.\par
\subsection{Output decoding}

As is seen in the \cite{mohapatra2021bevdetnet} architecture, the outputs are both classification and regression types. In the case of SNN, the outputs are mainly interpreted either by the number of spikes at the output neurons or by reading the membrane potential state of the output neurons. Since classification outputs are always discrete values, we read the number of spikes for keypoint and rotation output heads which are classification outputs. The neuron with the maximum number of spike outputs then gives the predicted class. For the regression output heads of box dimensions and center offsets, we read the membrane potentials of the output neurons for added precision.\par
\section{Experiments and initial results}

We trained the SNN model on the KITTI dataset using a 50/50 split and a 75/25 split for training and evaluation. We evaluated the trained model using an Intersection over Union (IoU) \cite{rezatofighi2019generalized} of 0.5 and 0.7, as seen in Table \ref{tab:my-table}. The spiking model performs very well when it has more training samples (using the 75/25 split). With the 50/50 split, the performance lags behind the CNN counterpart. This is partly attributed to the heavy quantizations that probably need more training samples to achieve a better generalization. Furthermore, we have removed an essential operation known as the Context Aggregation Module \cite{mohapatra2021bevdetnet} from the original CNN architecture since this did not have a proper replacement in the spiking domain. The model achieves similar inference time latency on a GPU but it is expected to achieve significantly lower power consumption on a custom SNN hardware. 
\section{Conclusion}

We have presented CNN to SNN conversion and simulation that adapts existing CNN building blocks to simulate spiking behavior for a complex real world application. We have shown that SNNs can be applied to complex tasks such as object detection and achieve comparable performance while achieving much better energy efficiency when implemented in hardware. This is a nascent research area, and we aim to progress from simulations to hardware implementation to realize the full potential of SNNs using other vital techniques such as event-based processing, differential signal processing, power-saving using sparsity, and low activation of neurons. 
\section*{Acknowledgment}
\begin{wrapfigure}{r}{0.04\textwidth}
\includegraphics[width=7mm]{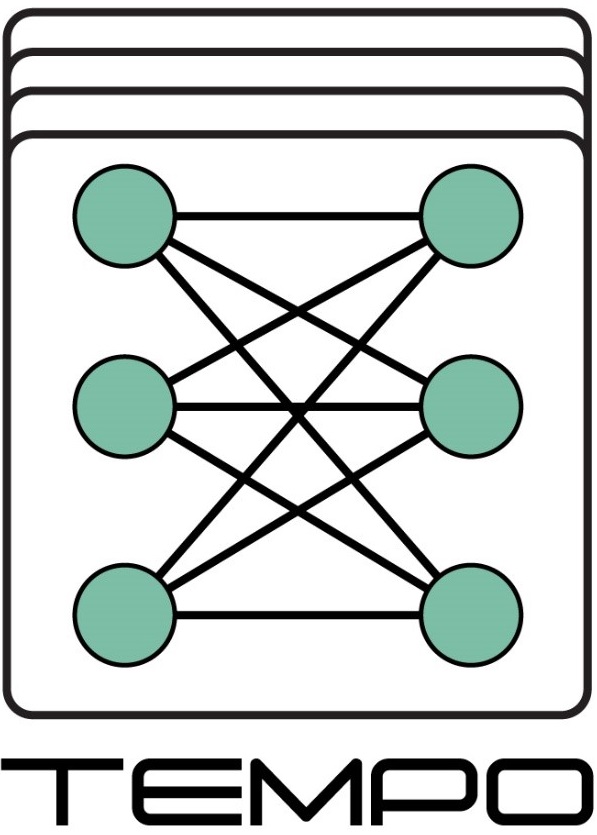}
\end{wrapfigure}
The Electronic Components and Systems fund us for European Leadership Joint Undertaking grant No 826655, receiving support from the European Union’s Horizon 2020 research and innovation program. Further partial funding is provided by the German Federal Ministry of Education and Research.
\bibliographystyle{IEEEtran}
\bibliography{references}
\end{document}